\pdfoutput=1
\documentclass{article}
\usepackage{setspace}
\usepackage{arxiv}
\usepackage[utf8]{inputenc} 
\usepackage[T1]{fontenc} 
\usepackage{booktabs} 
\usepackage{amsfonts} 
\usepackage{nicefrac} 
\usepackage{microtype} 
\usepackage{lipsum}
\usepackage{amsmath}
\usepackage{graphicx}
\usepackage{tabularx}
\usepackage{array}
\usepackage{multirow}
\usepackage{caption}
\DeclareMathOperator{\E}{\mathbb{E}}
\usepackage{hyperref}
\hypersetup{
 colorlinks=true,
 linkcolor = blue,
 filecolor = blue, 
 urlcolor = blue,
 citecolor = blue
}

\title{Estimating uncertainty of earthquake rupture using Bayesian neural network}

\author{
 Sabber Ahamed$^1$, Md Mesbah Uddin$^2$ \\
 \texttt{$^1$sabbers@gmail.com}, \texttt{$^2$Mesbahuddin1991@gmail.com} \\
}

\begin{document}
\maketitle

\begin{abstract}
    We introduce a new and practical artificial intelligence called a shallow explainable Bayesian neural network, specifically designed to tackle the challenges of limited data and high computational costs in earthquake rupture studies. Earthquake researchers often face the issue of insufficient data, which makes it difficult to determine the actual reasons behind an earthquake rupture. Furthermore, relying on trial-and-error simulations can be computationally expensive. We used 2,000 2D earthquake rupture simulations to train and test our model. Each simulation took about two hours on eight processors and had varying stress conditions and friction parameters. The BNN outperformed a standard neural network by 2.34$\%$. Upon analyzing the BNN's learned parameters, we found that normal stresses play a crucial role in determining rupture propagation and are the most uncertain factor, followed by the dynamic friction coefficient. Rupture propagation has more uncertainty than rupture arrest. Shear stress has a moderate impact, while geometric features like fault width and height are the least significant and uncertain factors. Our approach demonstrates the potential of the explainable Bayesian neural network in addressing the limited data and computational cost issues in earthquake rupture studies.
\end{abstract}


\section{Introduction}
Hazards due to earthquakes are a threat to economic losses and human life worldwide. A large earthquake produces high-intensity ground motion that causes damage to structures like buildings, bridges, and dams and takes many lives. Seismic hazard analysis (SHA) is the process used to estimate the risk associated with these damages. SHA requires historical earthquake information and detailed geological data. Unfortunately, there is a lack of a detailed surface and subsurface geologic information. Consequently, this leads to uncertainty in hazard estimation.

Numerical or physical models can be used to generate synthetic data that supplement existing data. Numerical models based studies such as dynamic earthquake rupture simulations need initial parameters about the fault and the surrounding region. Such parameters can include the stress-state, frictional, and material properties of the fault. However, the initial information is also not well constrained and not always available~\citep{Duan2006, peyrat2001dynamic, ripperger2008variability, kame2003effects}. Since earthquake rupture is a highly nonlinear process, determining the right initial parameter combinations is essential for the right simulation. Different initial conditions may lead to different results; therefore, we may not capture the real scenario of an actual earthquake rupture. The parameter combinations are often determined by making simplifying assumptions or taking a trial and error approach, which is computationally expensive~\citep{douilly20153d, ripperger2008variability, peyrat2001dynamic}. Therefore, high computational costs limit the applicability of simulations to integrate with seismic hazard analysis.

In recent years, machine learning (ML) approaches have been successfully used to solve many geophysical problems that have limited data and involve computational expense. For example, ~\cite{ahamed2019machine} developed a neural network and random forest algorithms to predict if an earthquake can break through a fault with geometric heterogeneity. The authors used 1600 simulated rupture data points for developing a neural network and random forest models. They were able to extract different patterns of the parameters that were responsible for earthquake rupture. The authors find that the models can predict rupture status within a fraction of a second. Average run time for a single simulation takes about two hours of wall-clock time on eight processors.

Machine learning approaches are also used in seismic event detection~\citep{RouetEartqhaukeMachine}, earthquake detection~\citep{perol2018convolutional}, identifying faults from unprocessed raw seismic data~\citep{last2016predicting} and to predict broadband earthquake ground motions from 3D physics-based numerical simulations~\citep{paolucci2018broadband}. All the examples show the potential of ML to solve many unsolved earthquake problems.

The performance of a machine learning model usually depends on the quality and the quantity of data. Bad quality or insufficient data may increase the uncertainty associated with each prediction~\citep{hoeting1999bayesian, blei2017variational, gal2017deep}. Prediction uncertainty estimation is vital in many applications: diseases detection~\citep{leibig2017leveraging, liu2018deep, nair2019exploring}, autonomous vehicle driving~\citep{kendall2015bayesian, mcallister2017concrete, burton2017making}, and estimating risk~\citep{hoeting1999bayesian, tong2001active, uusitalo2007advantages}. Therefore, calculating uncertainty is as crucial as improving model accuracy. All of the ML-based earthquake studies mentioned above avoid prediction uncertainty estimation.

To overcome the problem of insufficient data of earthquake rupture, we extend the work of~\citet{ahamed2019machine} using the Bayesian neural network. Unlike regular neural networks, BNN works better with a small amount of data and also provides prediction uncertainty. In this paper, we describe a workflow of (1) developing a BNN, (2) estimating prediction uncertainty, and (3) finding parameter combinations responsible for rupture. Identifying the source of uncertainty is vital to understanding the physics of earthquake rupture and estimating seismic risk. We also describe a technique that combines BNN and permutation importance to find the source of uncertainty.

\section{Rupture simulations and data processing}
\begin{figure}[ht]
    \begin{center}
        \includegraphics[scale=0.8]{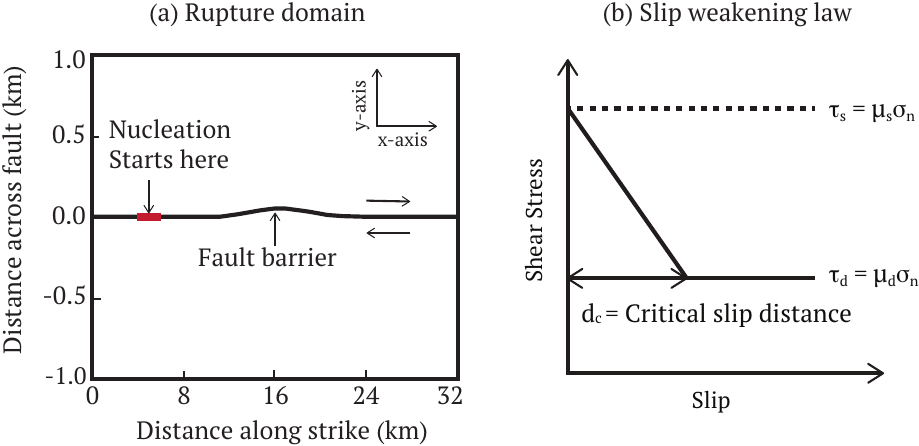}
    \end{center}
    \caption{(a) A zoomed view of the two-dimensional fault geometry. The domain is 32 km long along the strike of the fault and 24 kilometers wide across the fault. The rupture starts to nucleate 10 km to the left of the barrier and propagates from the hypocenter towards the barrier, (b) Linear slip-weakening friction law for an earthquake fault. The fault begins to slip when the shear stress reaches or exceeds the peak strength of $\tau_s$. $\tau_s$ decreases linearly with slip to a constant dynamic friction $\tau_d$ over critical slip distance ($d_c$). The shear strength is linearly proportional to the normal stress $\sigma_n$, and the friction coefficient varies with slip between $\mu_s$ and $\mu_d$.}
    \label{fig:rupture_domain}
\end{figure}

In this work, we used 200 simulated earthquake ruptures that were created by~\citet{ahamed2019machine}. The simulations are a two-dimensional rupture, illustrated in figure.~\ref{fig:rupture_domain}. The domain is 32 km long and 24 km wide. Figure~\ref{fig:rupture_domain}a shows the zoomed view of the original domain for better visualization of the fault barrier. An open-source \texttt{C++} and \texttt{python} based library $\texttt{fdfault}$~\citep{fdfaultEricGithub} was used to generate the ruptures. The library is a finite difference code for numerical simulation of elastodynamic fracture and friction problems. The code solves the elastodynamic wave equation coupled to a friction law describing the failure process. In each simulation, fault slip is calculated based on the initial stress conditions, the elastodynamic wave equations, and the frictional failure on that fault. The fault has a Gaussian geometric heterogeneity at the center. Rupture is nucleated 10 km to the left of the barrier and propagates towards the barrier. The linear slip-weakening law determines the strength of the fault (figure.~\ref{fig:rupture_domain}b). The fault starts to break when the shear stress ($\tau$) exceeds the peak strength $\tau_s = \mu_s \sigma_n$, where $\mu_s$ and $\sigma_n$ are the static friction coefficient and normal stress, respectively. Over a critical slip distance $d_c$, the friction coefficient reduces linearly to constant dynamic friction $\mu_d$. In each simulation, eight parameters were varied: x and y components of normal stress (sxx and syy), shear stress (sxy), dynamic friction coefficient, friction drop ($\mu_{s} - \mu_{d}$), critical slip distance ($d_c$), and width and height of the fault.

we used 1600 simulations to train the BNN. The remaining 400 were used to test the final model performance. The training dataset has an imbalance class proportion of rupture arrest (65$\%$) and rupture propagation (35$\%$). To avoid a bias toward rupture arrest, we upsampled the rupture propagation examples. Before training, all the data were normalized by subtracting the mean and dividing by the standard deviation.

\section{Neural network}
Neural networks (NN) are computing systems inspired by how neurons are connected in the brain~\citep{rosenblatt1958perceptron}. Several nodes are interconnected and organized in layers in a neural network. Each node is also known as a neuron. A layer can be connected to an arbitrary number of hidden layers of arbitrary size. However, increasing the number of hidden layers not always improve the performance but may force the model to generalize well on the training data but unseen (test) data, which is also known as overfitting~\citep{hinton2012improving, lawrence2000overfitting, lawrence1997lessons}. As a result, selecting the number of layers and nodes in each layer is one of the challenges of using neural networks in practice.

\section{Bayesian Neural network}
\begin{figure}[!h]
    \begin{center}
        \includegraphics[scale=0.90]{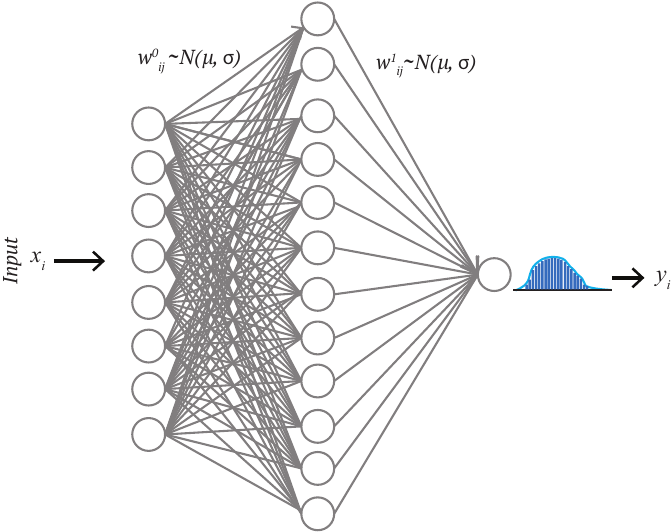}
    \end{center}
    \caption{The schematic diagram shows the architecture of the Bayesian neural network used in this work. The network has one input layer with eight parameters, one hidden layer with twelve nodes, and an output layer with a single node. Weights between input and hidden layers are defined by $w^0_{ij}$, which are normally distributed. $i, j$ are the node input and hidden layer node index. Similarly, $w^1_{jk}$ is the normal distribution of weights between the hidden and the output layer. $\mu$ and $\sigma$ are the mean and standard deviation. At the output node, the network produces a distribution of prediction scores between 0 and 1.}
    \label{fig:bnn-diagram}
\end{figure}

In a traditional neural network, weights are assigned as a single value or point estimate. In a BNN, weights are considered as a probability distribution. These probability distributions are used to estimate the uncertainty in weights and predictions. Figure~\ref{fig:bnn-diagram} shows a schematic diagram of a BNN where weights are normally distributed. The final learned network weights (posterior of the weights) are calculated using Bayes theorem as:

\begin{equation} \label{eq:bayes_formula}
    P(W|\textbf{X})= \frac{P(\textbf{X} \vert W) P(W)}{P(\textbf{X})}
\end{equation}

Where $\textbf{X}$ is the data, $P(\textbf{X} \vert W)$ is the likelihood of observing $\textbf{X}$, given weights ($W$). $P(W)$ is the prior belief of the weights, and the denominator $P(\textbf{X})$ is the probability of data which is also known as evidence. The equation requires integrating over all possible values of the weights as:

\begin{equation}
    P(\textbf{X}) = \int P(\textbf{X} \vert W) P(W) dW.
\end{equation}

Integrating over the indefinite weights in evidence makes it hard to find a closed-form analytical solution. As a result, simulation or numerical based alternative approaches such as Monte Carlo Markov chain (MCMC) and variational inference(VI) are considered. MCMC sampling is an inference method in modern Bayesian statistics, perhaps widely studied and applied in many situations. However, the technique is slow for large datasets and complex models. Variational inference (VI), on the other hand, is faster than MCMC. It has been applied to solve many large-scale computationally expensive neuroscience and computer vision problems~\citep{blei2017variational}.

In VI, a new distribution $Q(W \vert \theta)$ is considered that approximates the true posterior $P(W \vert \textbf{X})$. $Q(W \vert \theta)$ is parameterized by $\theta$ over $W$ and VI finds the right set of $\theta$ that minimizes the divergence of two distributions through optimization:

\begin{equation}\label{eq:optimization_kl}
    Q^{*}(W) = \operatorname*{argmin}_\theta \textbf{KL}\left[Q(W \vert \theta) || P(W|\textbf{X})\right]
\end{equation}

In equation-\ref{eq:optimization_kl}, \textbf{KL} or Kullback–Leibler divergence is a non-symmetric and information theoretic measure of similarity (relative entropy) between true and approximated distributions~\citep{kullback1997information}. The KL-divergence between $Q(W \vert \theta)$ and $P(W|\textbf{X})$ is defined as:

\begin{equation}
    \textbf{KL}\left[Q(W \vert \theta) || P(W|\textbf{X})\right] = \int Q(W \vert \theta)\log\frac{Q(W \vert \theta)}{P(W \vert \textbf{X})}dW
\end{equation}

Replacing $P(W|\textbf{X})$ using equation-\ref{eq:bayes_formula} we get:

\begin{align}
    \textbf{KL}\left[Q(W \vert \theta) || P(W|\textbf{X})\right] & = \int Q(W \vert \theta)\log\frac{Q(W \vert \theta)P(\textbf{X})}{P(\textbf{X} \vert W) P(W)}dW                                                                 \\
                                                                 & = \int Q(W \vert \theta)\left[\log Q(W \vert \theta)P(\textbf{X}) - \log P(\textbf{X} \vert W) P(W)\right] dW                                                   \\
                                                                 & = \int Q(W \vert \theta)\log \frac{Q(W \vert \theta)}{P(W)}dW + \int Q(W \vert \theta)\log P(\textbf{X})dW - \int Q(W \vert \theta)\log P(\textbf{X} \vert W)dW
\end{align}

Taking the expectation with respect to $Q(W \vert \theta)$, we get:

\begin{align}
    \textbf{KL}\left[Q(W \vert \theta) || P(W|\textbf{X})\right]
     & = \E\left[\log \frac{Q(W \vert \theta)}{P(W)}\right] + \log P(\textbf{X})- \E\left[\log P(\textbf{X} \vert W)\right]
\end{align}

The above equation shows the dependency of $\log P(\textbf{X})$ that makes it difficult to compute. An alternative objective function is therefore, derived by adding $\log P(\textbf{X})$ with negative KL divergence. $\log P(\textbf{X})$ is a constant with respect to $Q(W \vert \theta)$. The new function is called as the evidence of lower bound (ELBO) and expressed as:

\begin{align}
    ELBO(Q) & = \E\left[\log P(\textbf{X} \vert W)\right] - \E\left[\log \frac{Q(W \vert \theta)}{P(W)}\right]  \\
            & = \E\left[\log P(\textbf{X} \vert W)\right] - KL\left[Q(W \vert \theta) || P(W|\textbf{X})\right]
\end{align}

The first term is called likelihood, and the second term is the negative KL divergence between a variational distribution and prior weight distribution. Therefore, ELBO balances between the likelihood and the prior. The ELBO objective function can be optimized to minimize the KL divergence using different optimizing algorithms like gradient descent.

\section{Train the BNN}
The BNN has the same NN architecture used in~\citet{ahamed2019machine} to compare the performance between them. Whether the BNN performs better or similarly to NN, BNN provides an additional advantage of prediction uncertainty. Like NN, BNN has one input layer with eight parameters, one hidden layer with twelve nodes and one output layer (Figure~\ref{fig:bnn-diagram}). A nonlinear activation function $\texttt{ReLu}$~\citep{hahnloser2000digital} was used at the hidden layer. $\texttt{ReLu}$ passes all the values greater than zero and sets the negative output to zero. The output layer uses \texttt{sigmoid} activation function, which converts the outputs between zero and one.

Prior weights and biases are normally distributed with zero mean and one standard deviation. Figure~\ref{fig:posterioir_weight} shows the log density of prior and posterior weights ($w^k_{ij}$) and biases ($b^k_j$). $i$ and $j$ are the index of the input and hidden layer nodes. $i$ ranges from 0 to 7, and $j$ ranges from 0 to 11. $k$ is the index that maps two layers. As an example, $w^0_{15}$ is the weight between the first input node and the fifth hidden node. The output node of the last layer produces a distribution of prediction scores between 0 and 1. The prediction distributions are used to compute standard deviation, which is the uncertainty metric.

Adam optimization (extension of stochastic gradient descent) was used to minimize the KL divergence by finding a suitable variational parameter $\theta$. The initial learning rate is 0.5, which exponentially decays as the training progresses. To train the BNN, we use \texttt{Edward}~\citep{tran2016edward, tran2017deep}, \texttt{TensoFlow}~\citep{tensorflow2015-whitepaper} and \texttt{Scikit-learn}~\citep{pedregosa2011scikit}. \texttt{Edward} is a Python-based Bayesian deep learning library for probabilistic modeling, inference, and criticism. All the training data, codes, and the corresponding visualizations can be found on the Github repository: \url{https://github.com/msahamed/earthquake_physics_bayesian_nn}

\subsection{Prior and posterioir weight distribution}
\begin{figure}[!h]
    \begin{center}
        \includegraphics[scale=0.50]{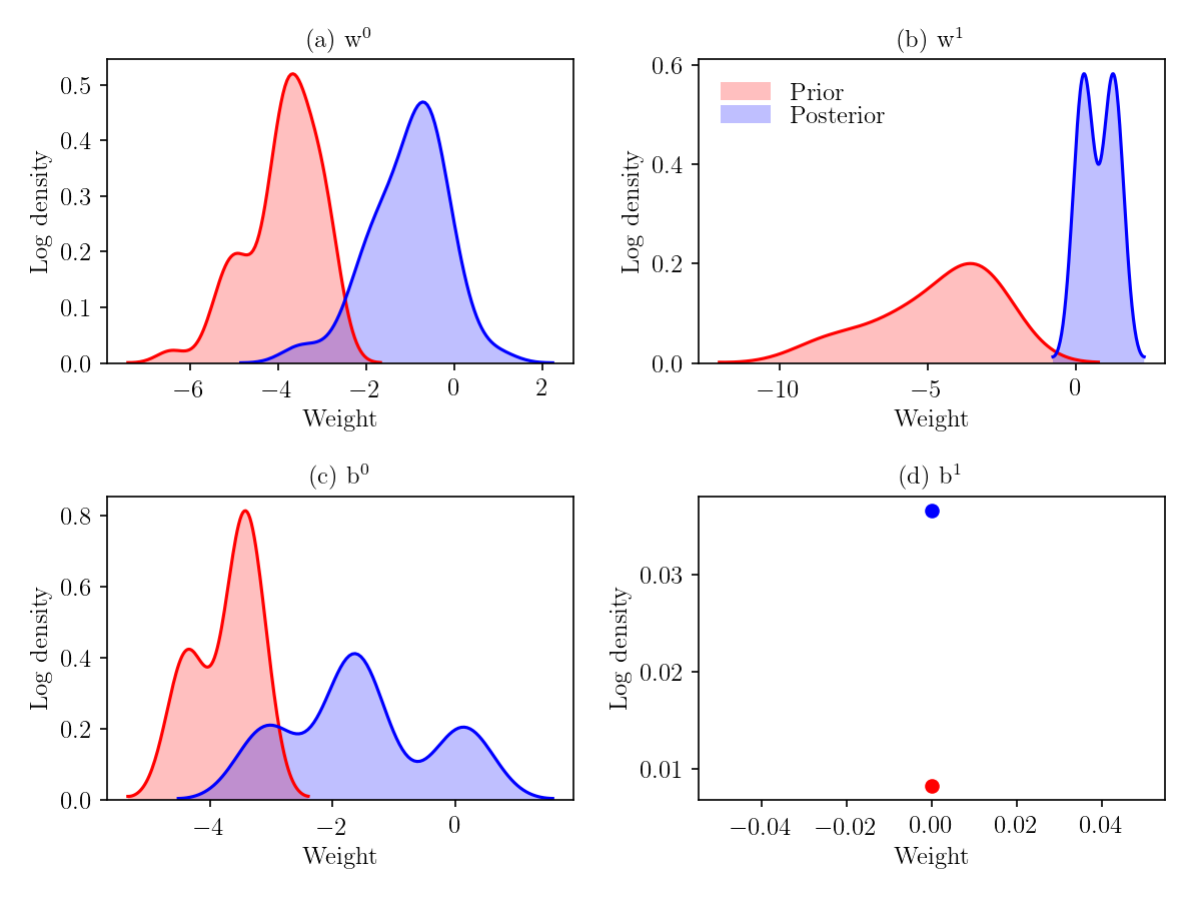}
    \end{center}
    \caption{The graph shows the distribution of prior and posterior mean weights (a) $w^0$ (b) $w^1$ and biases (c) $b^0$ (d) $b^1$. Both location of the mean and magnitude of density of the posterior distributions (weights and biases) are noticeably different from the priors which indicates that BNN has learned from the data and adjusted the posterior accordingly.}
    \label{fig:posterioir_weight}
\end{figure}

To evaluate the parameters (weights and biases) of the BNN, 1000 posterior samples of $w^0_{ij}$, $w^1_{jk}$, $b^0$ and $b^1$ were used. Figure~\ref{fig:posterioir_weight} shows the prior and posterior distribution of mean weight and biases. The posterior location of the mean and density of the weights and biases are different from their priors. For example, the location of $w^0$ shifts toward non-negative value, while the density remains similar. Whereas, the $w^1$, $b^0$, and $b^1$ have a different posterior mean location and density than their prior. The differences between prior and posterior indicate that the BNN has learned from the data and adjusted the posterior distribution accordingly.

\subsection{BNN classification result}
The performance of the BNN was evaluated using 400 test simulations. For a given test example, 1000 posterior samples of the weights and biases produce 1000 prediction scores. The scores were then used to determine the prediction class and associated uncertainty (standard deviation). To determine the proper class of the examples, the mean score was calculated from the 1000 predictions for each example. Then an optimal threshold was computed that maximizes the model F-1 score (0.54). If the mean prediction score of an example was greater or equal to the optimal threshold, then the earthquake was classified as propagated, and otherwise, arrested. Uncertainty is the standard deviation of the prediction scores. The F-1 score is the harmonic mean of the true positive rate and precision of the model. Table~\ref{tab:nn_confusion matrix} shows the confusion matrix for the actual and predicted classifications. The test accuracy of the BNN is $83.34\%$, which is $2.34\%$ higher than NN. BNN also reduces the four false positives (FP) and three false negatives (FN). Table~\ref{tab:nn_classification_report} shows the detailed classification report of the model performance. The results imply that BNN has the potential to improve performance. Since BNN produces distributions of the score rather than a point estimation, BNN can better generalize the unseen data and thus help reduce overfitting.

\begin{table}[!h]
    \caption{Confusion matrix of 400 test data shows the performance of the BNN}
    \label{tab:nn_confusion matrix}
    \centering
    \begin{tabular}{l c c}
        \hline
                             & Actual propagated & Actual arrested \\
        \hline
        Predicted propagated & 226               & 46              \\
        Predicted arrested   & 22                & 106             \\
        \hline
    \end{tabular}
\end{table}

\begin{table}[h]
    \caption{Classification resultsof 400 test data}
    \label{tab:nn_classification_report}
    \centering
    \begin{tabular}{l c c c c}
        \hline
        Class              & Precision & Recall & F1 score & support \\
        \hline
        Rupture arrested   & 0.91      & 0.83   & 0.87     & 272     \\
        Rupture propagated & 0.70      & 0.83   & 0.76     & 128     \\
        Average/Total      & 0.84      & 0.83   & 0.83     & 400     \\[1ex]
        \hline
    \end{tabular}
\end{table}

\section{Uncertainity analysis}
Uncertainty analysis helps to make better decisions and estimate risk accurately. In the following subsection, we discuss three types of uncertainties: 1. network uncertainty, 2. prediction uncertainty and 3. feature uncertainty, which was estimated from BNN and their implication on the earthquake rupture. We also discuss how uncertainty can help us understand physics and find the parameter combinations responsible for an earthquake rupture.

\subsection{Network uncertainity}

\begin{figure}[!h]
    \begin{center}
        \includegraphics[scale=0.550]{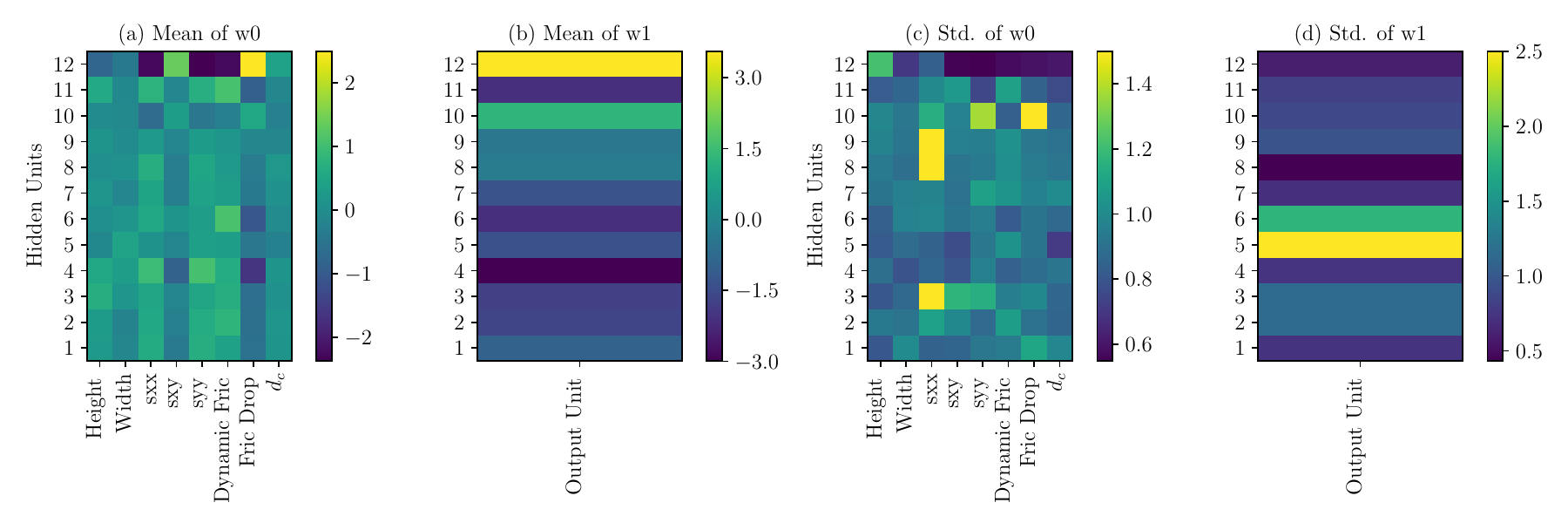}
    \end{center}
    \caption{The illustration shows the posterior mean and standard deviations of $w^0$ and $w^1$. (a) $w^0$ that map the inputs to the nodes of the hidden layer. The eight input parameters are on the horizontal axis, and the twelve nodes are on the vertical axis. The colors in each cell are the magnitudes of mean weight. (b) $w^1$ maps the hidden layer to the output layer. (c) The standard deviation of $w^0$. Shear stress connected to node-4 of the hidden layer has the highest uncertainty. Similarly, the weights associated with the input parameters and node-5 have high uncertainty. Whereas, the weights associated with the input parameters and the nodes 7-11 have relatively low uncertainty. (d) Uncertainty of the weights associated with the hidden layer nodes and the output node. Weights in node 7 and 8 have high uncertainty while the rest of the weights have relatively low uncertainty.}
    \label{fig:network_uncertainity}
\end{figure}

Estimating the uncertainty of neural network parameters (weights) helps us understand the black box behavior. The illustration in Figure~\ref{fig:network_uncertainity} shows the mean and standard deviation of $W^0$ and $W^1$. $W^0$ maps the inputs to the hidden layer nodes, whereas $W^1$ maps the output node of the output layer to the nodes of the hidden layer. The colors in each cell indicate the magnitude of a weight that connects two nodes. In the input and hidden layer, the ReLu activation function was used. ReLu passes all the positive output while setting a negative output to zero. At the output layer, sigmoid activation is used, which pushes the larger weights toward one and smaller or negative weights toward zero. Therefore, positive and high magnitude weights contribute to the earthquake rupture and vice versa.

In $w^0$, nodes connected to friction drop, dynamic friction, shear, and normal stresses have variable positive and negative weights. The corresponding node in $w^1$ also has a strong positive or negative magnitude. For example, node-12 has both positive and negative weights in $w^0$, and the corresponding node in $w^1$ has a high positive weight. Similarly, node-4 has a substantial negative weight in $w^1$, and the corresponding nodes in $w^0$ have both positive and negative weights. On the other hand, width, height, and $d_c$ have a similar magnitude of weights, and the corresponding nodes in $w^1$ have a moderate magnitude of weights. Thus the variable weights make friction drop, dynamic friction, shear, and normal stresses influential on the prediction score. Therefore, for any combined patterns of the input features, we can now detect the critical features and the source uncertainties.

For example, node-10 of $w^1$ has positive weight (Figure~\ref{fig:network_uncertainity}b). In $w^0$, the corresponding connecting input features, friction drop, and shear stress have positive weight and low uncertainty, whereas the rest of the features have a similar magnitude of weight. The combination of high friction drop and shear stress weight and low weights of other features increase the prediction score, thus likely to cause rupture to propagate. Friction drop and normal stresses also have high uncertainty (Figure~\ref{fig:network_uncertainity}c). For this combination of patterns, friction drop and normal stresses influence the prediction strongly and are also the sources of uncertainty. Thus, it gives us the ability to investigate any rupture propagation example in terms of uncertainty.

\subsection{Prediction uncertainity}

\begin{figure}[!h]
    \begin{center}
        \includegraphics[scale=0.50]{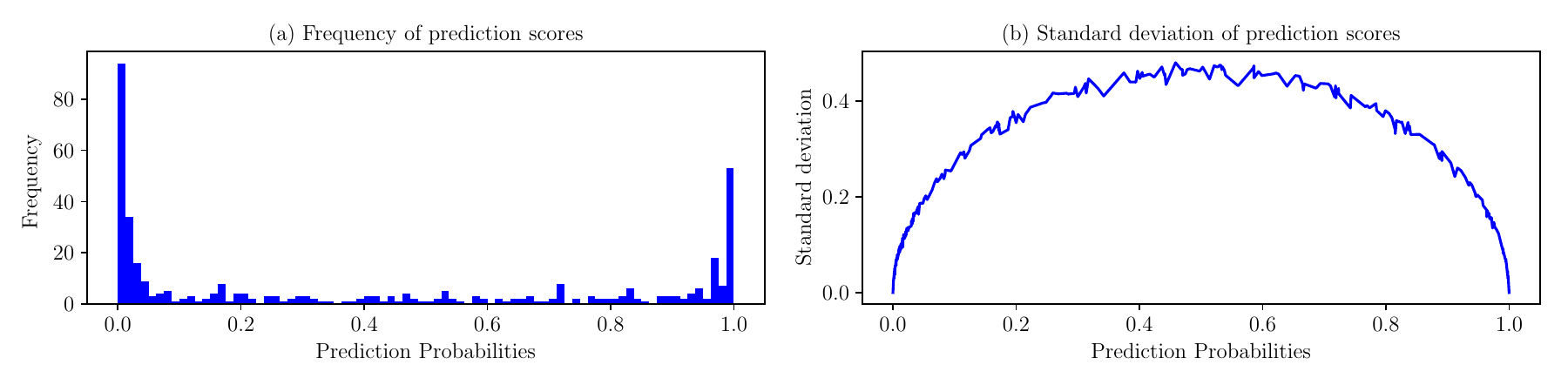}
    \end{center}
    \caption{The graph shows (a) frequency and (b) standard deviation of posterior prediction scores of the test data. Prediction scores are skewed toward the left side while slightly less on the right. The observation is consistent with the proportion of the rupture arrest (272) and rupture propagation (120) in the test data. Prediction scores close to zero are related to rupture arrest, and scores around one are the rupture propagation. Standard deviations are high with scores around 0.5.}
    \label{fig:posterioir_uncertainity_statistics}
\end{figure}

Figure~\ref{fig:posterioir_uncertainity_statistics}a and b show the frequency and standard deviation of test data prediction scores. The distribution is more skewed toward smaller scores than the higher ones. Scores close to zero are associated with rupture arrest, whereas scores close to one are rupture propagations. The observation is consistent with the class proportion of the test data. Rupture arrest has a higher number of examples (272) than rupture propagation (120). Figure~\ref{fig:posterioir_uncertainity_statistics}a shows that scores roughly between 0.35 and 0.75 have a fewer number of examples and thus high uncertainty. The likely reason for the high uncertainty is that the examples have both rupture propagation and arrest properties. As a result, the model gets confused and cannot classify the example correctly; thus, the misclassification rate is high in this region. The performance of the network could be improved if sufficient similar example data are added to the training dataset.

\subsection{Feature uncertainity}
\begin{figure}[!h]
    \begin{center}
        \includegraphics[scale=0.650]{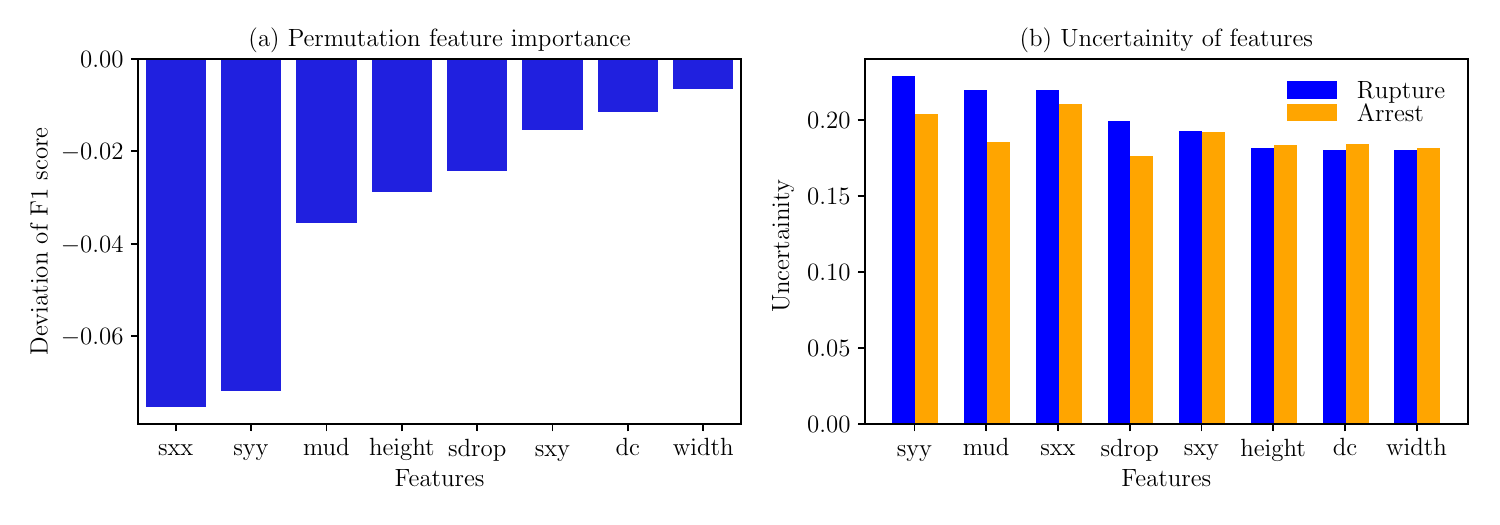}
    \end{center}
    \caption{The illustration shows (a) permutation feature importance and (b) their uncertanitities. }
    \label{fig:permutation_importance_uncertainity}
\end{figure}

We used the permutation importance method to determine the source of uncertainty in the test data. Permutation importance is a model agnostic method that measures the influencing capacity of a feature by shuffling it and measuring corresponding global performance deviation. If a feature is a good predictor, then altering its values reduces the model’s global performance significantly. The shuffled feature with the highest performance deviation is the most important and vice versa. In this work, the F-1 score is the performance measuring metric.

Figure~\ref{fig:permutation_importance_uncertainity}(a) shows the permutation importance of all the features. Normal stresses (sxx and syy) have the highest F-1 score deviation, which is approximate $~7\%$ less than the base performance, followed by the dynamic friction coefficient. Geometric feature width has the least contribution role to determine the earthquake rupture. These observations are consistent with the observation of~\citet{ahamed2019machine}, where the authors rank the features based on the random forest feature importance algorithm.

From the distribution of prediction scores, the standard deviation was calculated for each shuffled feature. Figure~\ref{fig:permutation_importance_uncertainity}(b) shows the uncertainty of each feature of earthquake rupture propagation and arrest. All features have higher uncertainty in the rupture propagation compare to the rupture arrest. In both classes, the major portion of uncertainty comes from normal stresses. Shear stress, height, width, and critical distance ($d_c$) have a similar amount of uncertainty in both of the classes. The dynamic friction coefficient and friction drop are also the comparable uncertainty sources in rupture to propagate while it is slightly less in earthquake arrest.

The above observations imply that normal stress and friction parameters have a more considerable influence in determining the earthquake rupture. Although the height and width of a fault are not a significant source of uncertainty, they play a more influential role in influencing other features. For example, in a complex rough fault, the variation of the bending angle of barriers affects stress perturbation, consequently increasing the uncertainty. If the angle near the bend is sharp, the variation in traction at the releasing and restraining bend is more prominent. Whereas if the barrier is broad, the stress perturbation at the restraining and releasing bend is less noticeable. ~\citet{chester2000stress} found a similar observation that fault geometry impacts the orientation and magnitude of principal stress.

\section{Discussion and conclusion}
In recent years, deep learning has been used to solve many real-life problems and achieve a state of the art performance. For example, facial recognition, language translation, self-driving cars, disease identification, and more. Such successful applications require millions of data points to train the model and achieve a state of the art performance. However, many problem spaces have limited data. This has been a barrier to the use of deep learning to solve many other real-life problems, like earthquake rupture. A Bayesian neural network, on the other hand, can perform well and achieve excellent performance even on small datasets.

In this work, we use a Bayesian neural network to combat the small data problem and estimate the uncertainty for earthquake rupture. Two thousand rupture simulations generated by~\citet{ahamed2019machine} were used to train and test the model. Each 2D simulation fault has a Gaussian geometric heterogeneity at the center. The eight fault parameters of normal shear stress, height, and width of the fault, stress drop, dynamic friction coefficient, critical distance were varied in each of the simulations. Sixteen hundred simulations were used to train the BNN, and 400 were used to test the generalization of the model. The BNN has the same architecture as the NN of ~\citet{ahamed2019machine}.

The test F1 score is 0.8334, which is 2.34$\%$ higher than the NN. All the features have a higher uncertainty in rupture propagation than the rupture arrest. The highest sources of uncertainty came from normal stresses, followed by a dynamic friction coefficient. Therefore, these features have a higher influencing capacity in determining the prediction score. Shear stress has a moderate role, but the geometric features such as the width and height of the fault are least significant in determining rupture. Test examples with prediction scores around 0.5 have a higher uncertainty than those with low (0-0.30) and high (0.7-1.0) prediction scores. Cases with prediction scores around 0.5 have mixed properties of rupture propagation and arrest.

\bibliographystyle{unsrtnat}
\bibliography{bayesian_nn_earthquake}

\end{document}